\def\year{2022}\relax
\documentclass[letterpaper]{article} 
\usepackage{aaai22}  
\usepackage{times}  
\usepackage{helvet}  
\usepackage{courier}  
\usepackage[hyphens]{url}  
\usepackage{graphicx} 
\usepackage{natbib}  
\usepackage{caption} 
\DeclareCaptionStyle{ruled}{labelfont=normalfont,labelsep=colon,strut=off} 
\usepackage{algorithm}
\usepackage{algorithmic}
\usepackage{booktabs}
\usepackage{amsfonts}
\usepackage{multirow}
\usepackage{amsmath}

\newcommand{\tabincell}[2]{\begin{tabular}{@{}#1@{}}#2\end{tabular}}

\usepackage{newfloat}
\usepackage{listings}
\floatstyle{ruled}
\newfloat{listing}{tb}{lst}{}
\floatname{listing}{Listing}
\title{Deep Neural Networks Learn Meta-Structures from Noisy Labels in Semantic Segmentation}
\author{
	 Yaoru Luo \textsuperscript{\rm 1, \rm 2},
	 Guole Liu \textsuperscript{\rm 1, \rm 2},
	 Yuanhao Guo \textsuperscript{\rm 1, \rm 2},
	 Ge Yang \textsuperscript{\rm 1, \rm 2} \thanks{Corresponding Author.}
}
\affiliations{

	\textsuperscript{\rm 1}National Laboratory of Pattern Recognition, Institute of Automation, Chinese Academy of Sciences\\
    \textsuperscript{\rm 2}School of Artificial Intelligence, University of Chinese Academy of Sciences\\
    \{luoyaoru2019, yuanhao.guo\}@ia.ac.cn, \{liuguole, yangge\}@ucas.edu.cn
    
}

\begin{document}

\maketitle

\begin{abstract}

How deep neural networks (DNNs) learn from noisy labels has been studied extensively in image classification but much less in image segmentation. So far, our understanding of the learning behavior of DNNs trained by noisy segmentation labels remains limited. In this study, we address this deficiency in both binary segmentation of biological microscopy images and multi-class segmentation of natural images. We generate extremely noisy labels by randomly sampling a small fraction (e.g., 10\%) or flipping a large fraction (e.g., 90\%) of the ground truth labels.	When trained with these noisy labels, DNNs provide largely the same segmentation performance as trained by the original ground truth. \textit{This indicates that DNNs learn structures hidden in labels rather than pixel-level labels per se in their supervised training for semantic segmentation}. We refer to these hidden structures in labels as meta-structures. When DNNs are trained by labels with different perturbations to the meta-structure, we find consistent degradation in their segmentation performance. In contrast, incorporation of meta-structure information substantially improves performance of an unsupervised segmentation model developed for binary semantic segmentation. We define meta-structures mathematically as spatial density distributions and show both theoretically and experimentally how this formulation explains key observed learning behavior of DNNs.

\end{abstract}

\section{Introduction}
Deep neural networks (DNNs) have shown excellent performance in challenging image segmentation tasks \cite{long2015fully, DBLP:journals/corr/YuK15, jegou2017one, chen2017deeplab}. But their supervised training requires pixel labels for each training image. Manual annotation of pixels not only is laborious but also easily introduces label noise, especially in border regions. Compared to image classification, label noise is more common in image segmentation. So far, however, studies on the learning behavior of DNNs trained by noisy labels have focused primarily on classification. Is it necessary to accurately label each pixel for accurate segmentation? How is segmentation performance of DNNs influenced when trained by different types of noisy labels? Answering these questions not only will help us understand the role of labels in training of DNNs but also will provide insights into the learning behavior of DNNs.



\begin{table}[t]
\centering
\begin{tabular}{lll}
\toprule
	Abbreviation & Meaning & Description \\
\midrule
    CL      &  Clean Label       & \tabincell{l}{Ground truth labels \\ annotated by human \\ experts} \\ 
\cline{1-3}
    RCL    & \tabincell{l}{Randomized \\ Clean Label}       & \tabincell{l}{Randomly sampled \\ or flipped pixel labels \\ from ground truth}  \\

\cline{1-3}
    PCL 	   & \tabincell{l}{Perturbed \\ Clean Label} & \tabincell{l}{Dilation/Erosion/\\ Skeleton of ground \\ truth} \\
\cline{1-3}
    RL         & Random Label      & \tabincell{l}{Randomly generated \\ pixel labels}    \\
\bottomrule
\end{tabular}
\caption{Different types of labels used in this study.}
\label{table1}
\end{table}

Different types of noisy labels have been used to elucidate the learning behaviors of DNNs in image classification, including partially corrupted or randomly shuffled labels \cite{DBLP:conf/iclr/ZhangBHRV17, DBLP:conf/icml/ArpitJBKBKMFCBL17}. In image segmentation, there are image-level and pixel-level label noises. Image-level label noise refer to erroneous semantic annotation of image objects. Pixel-level label noise refer to erroneous semantic annotation of image pixels. In this study, we focus on pixel-level label noise.  

We examine the performance of DNNs trained by four different types of labels, as summarized in Table 1 and shown in Figure 1 and Figure 4. To quantitatively analyze the segmentation performance of DNNs trained by these labels, we experiment on two representative segmentation models, U-Net \cite{ronneberger2015u} and DeepLabv3+ \cite{chen2018encoder}, with the same loss function (binary cross-entropy loss) and optimizer (stochastic gradient descent, SGD). Performance of DNNs trained by the labels ranks from the best to the worst as follows:
\begin{equation}
	CL \approx RCL > PCL > RL
\end{equation}
As shown in Figure 1, when U-Net is trained with 45\% of the labels randomly flipped (RCL) in binary segmentation, its performance remains largely the same as trained by the original ground truth (CL). Similar results on DeepLabv3+ are shown in Appendix A. These results indicate that DNNs learn structures hidden in the noisy labels rather than the pixel labels per se in their training for segmentation. We refer to these hidden structures as meta-structures. 

\begin{figure}[t]
\centering
\includegraphics[width=0.9\columnwidth]{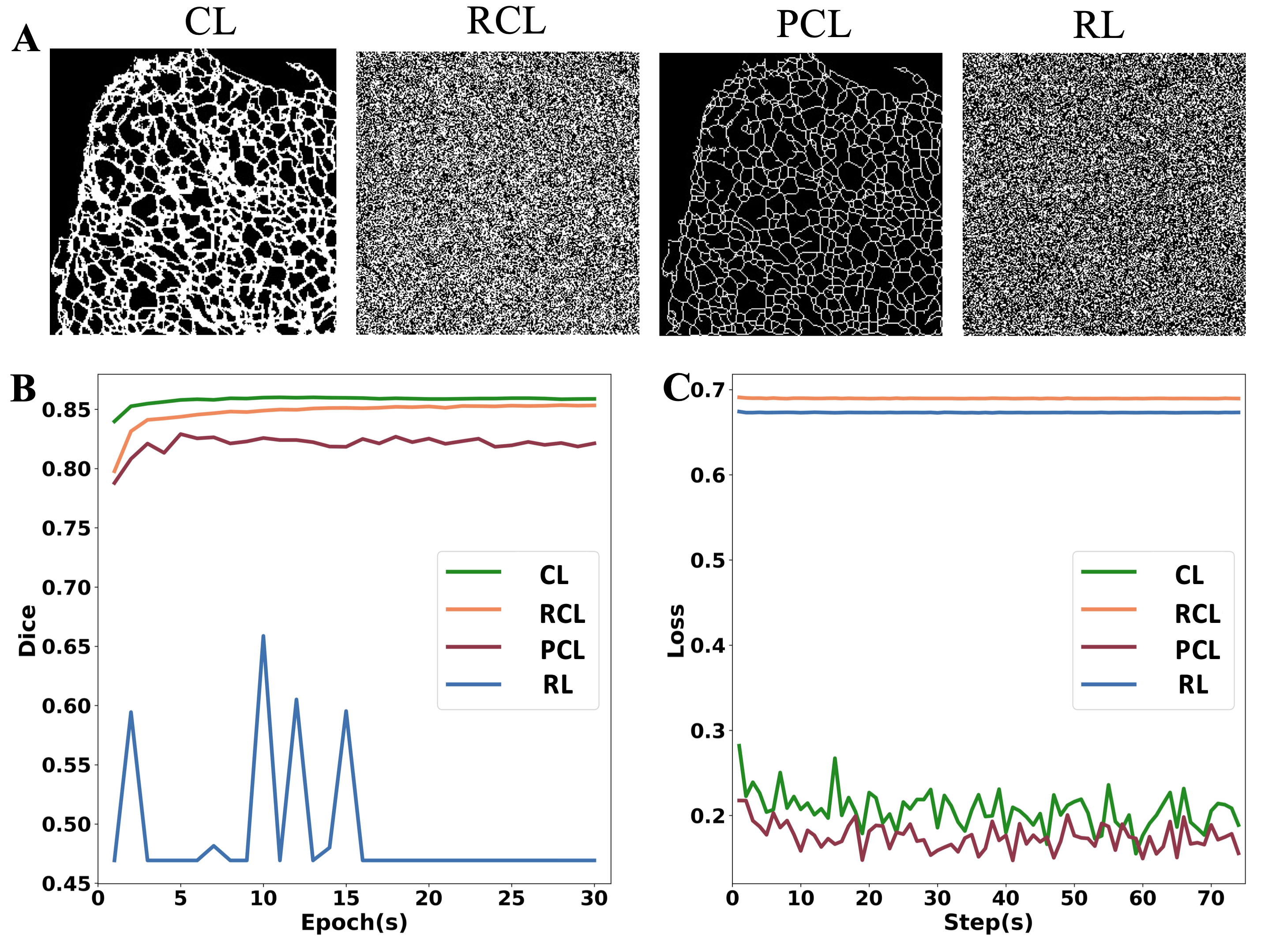} 
\caption{Segmentation performance of DNNs trained by different types of labels. (A) Four types of training labels for an image of the endoplasmic reticulum from the ER dataset. CL: ground truth from manual annotation. RCL: each pixel label in CL is randomly flipped with a probability of 0.45. PCL: Skeleton of CL. RL: each pixel is randomly annotated as 1, i.e. foreground, with a probability of 0.5. (B) Testing dice scores during training. (C) Training loss of each batch optimization step.}
\label{fig1}
\end{figure}

Similar as observed in image classification in \cite{DBLP:conf/iclr/ZhangBHRV17}, we also find that DNNs memorize random labels in segmentation since the training loss under RCL and RL quickly converges to a constant but not under CL and PCL (Figure 1C). Meanwhile, similar as observed in image classification in \cite{DBLP:conf/icml/ArpitJBKBKMFCBL17}, we also find that before memorizing RL, DNNs prioritize learning real patterns first in segmentation because the dice score (Figure 1B, blue line) first fluctuates greatly then quickly drops to a low level. Since RL requires no annotation, this motivates us to develop an unsupervised segmentation model for binary segmentation. The model sets RL as the initial training label and iteratively updates the training label to RCL by incorporating meta-structure information.

\noindent \textbf{Main Contributions}

\noindent The main research contributions of this study are as follows:

\begin{itemize}
\item We provide direct experimental evidence that DNNs learn implicit structures hidden in noisy labels in semantic segmentation. We name these implicit structures as meta-structures and model them mathematically as spatial density distributions. We show theoretically and experimentally how this model may quantify semantic meta-structure information in the noisy labels.
\item We have identified some basic properties of meta-structures. We find that DNNs trained with labels under different perturbations to the meta-structures exhibit consistently worse segmentation performance.
\item By incorporating meta-structure information, we have developed an unsupervised model for binary segmentation that outperforms the state-of-the-art unsupervised models and achieves remarkably competitive performance against supervised models.
\end{itemize}

\section{Methods}

%

\subsection{Generation of different Noisy Labels}
We synthesize different types of noisy labels as follows:

\textbf{(1) RCL:} We synthesize randomized clean label (RCL) by randomly sampling or flipping pixel labels in CL. For random sampling, we randomly select pixel labels of each class with a probability of $P_{sample}$ and exclude unsampled pixels from training. For random flipping, we randomly swapped a fraction of $P_{flip}$ of true labels with randomly selected labels from other classes. See Figure 2 and Figure 3 for examples of randomized binary labels and multi-class labels, respectively. 

\textbf{(2) PCL:} We perform image dilation or erosion with a $3 \times 3$ template or extract one-pixel-wide skeleton of CL. We refer to these types of noisy labels as perturbed clean labels (PCL). Examples of PCL are shown in Figure 4A.

\textbf{(3) RL:} We synthesize random label (RL) by randomly annotating pixel labels with a probability $P_{generate}$. RL can be considered as a strong perturbation of CL since it contains no information from CL. Examples of RL are shown in Figure 4B.

\subsection{Experiment Configuration}
We examine the learning behavior of two DNNs (U-Net and DeepLabv3+) in both binary-class and multi-class segmentation. For binary-class segmentation, we select fluorescence microscopy images of ER, MITO datasets \cite{t2he-zn97-20} and the NUC dataset \cite{caicedo2019nucleus}. For multi-class segmentation, we select natural images of Cityscapes dataset \cite{cordts2016cityscapes}. Detailed information on the datasets and experimental configurations are provided in Appendix C.


\section{Experimental Evidence for Existence of Meta-Structures}

We examine the learning behavior of DNNs trained by RCL in both binary-class and multi-class segmentation.

\subsection{Existence of Meta-Structures}

Figure 2 shows the testing dice scores of U-Net (Solid lines) and DeepLabv3+ (Dashed line) trained by binary RCL. Although the rate of convergence generally decreases under increase of label randomization, especially for DeepLabv3+, the final dice scores decrease only slightly and the maximum gap between final dice scores of CL and RCL is $7\%$. Similar results are observed on other binary datasets. See Appendix C.3 for details.

\begin{figure}[t]
\centering
\includegraphics[width=0.9\columnwidth]{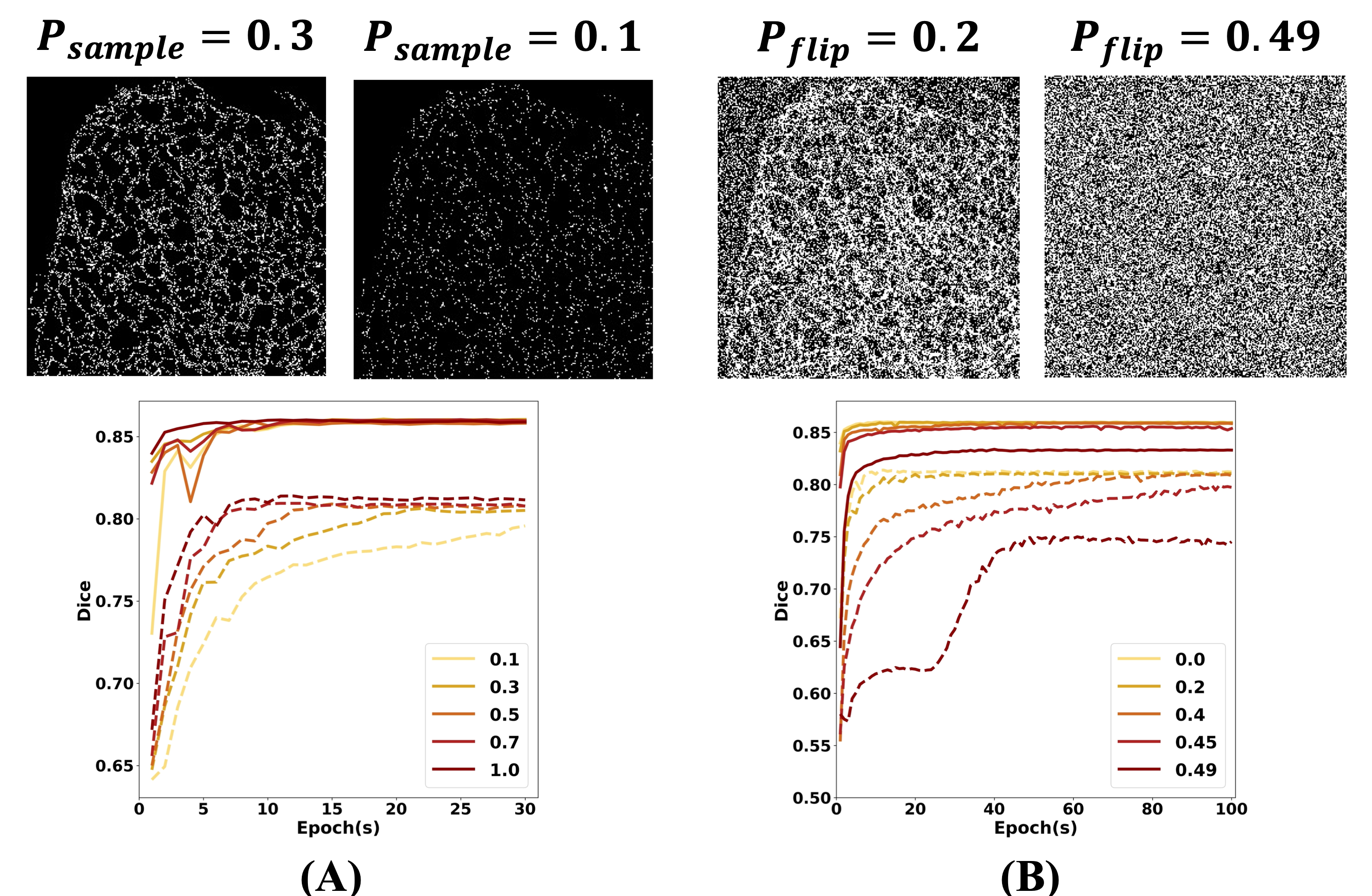} 
\caption{Segmentation performance of DNNs trained by RCL generated by (A) random sampling and (B) random flipping on ER dataset. Solid lines: U-Net. Dashed lines: DeepLabv3+.}
\label{fig2}
\end{figure}

For multi-class labels, we train DeepLabv3 on the natural image dataset \textit{Cityscapes} and compare with results of a previous study as our baseline \cite{chen2017rethinking}. Due to a lack of previous results for comparison, we did not experiment with U-Net. Representative results are shown in Figure 3. Table 2 summarizes segmentation results of RCLs under different randomization probabilities. No degradation in segmentation performance measured in mean intersection over union (mIoU) is observed.

\begin{figure}[t]
\centering
\includegraphics[width=0.9\columnwidth]{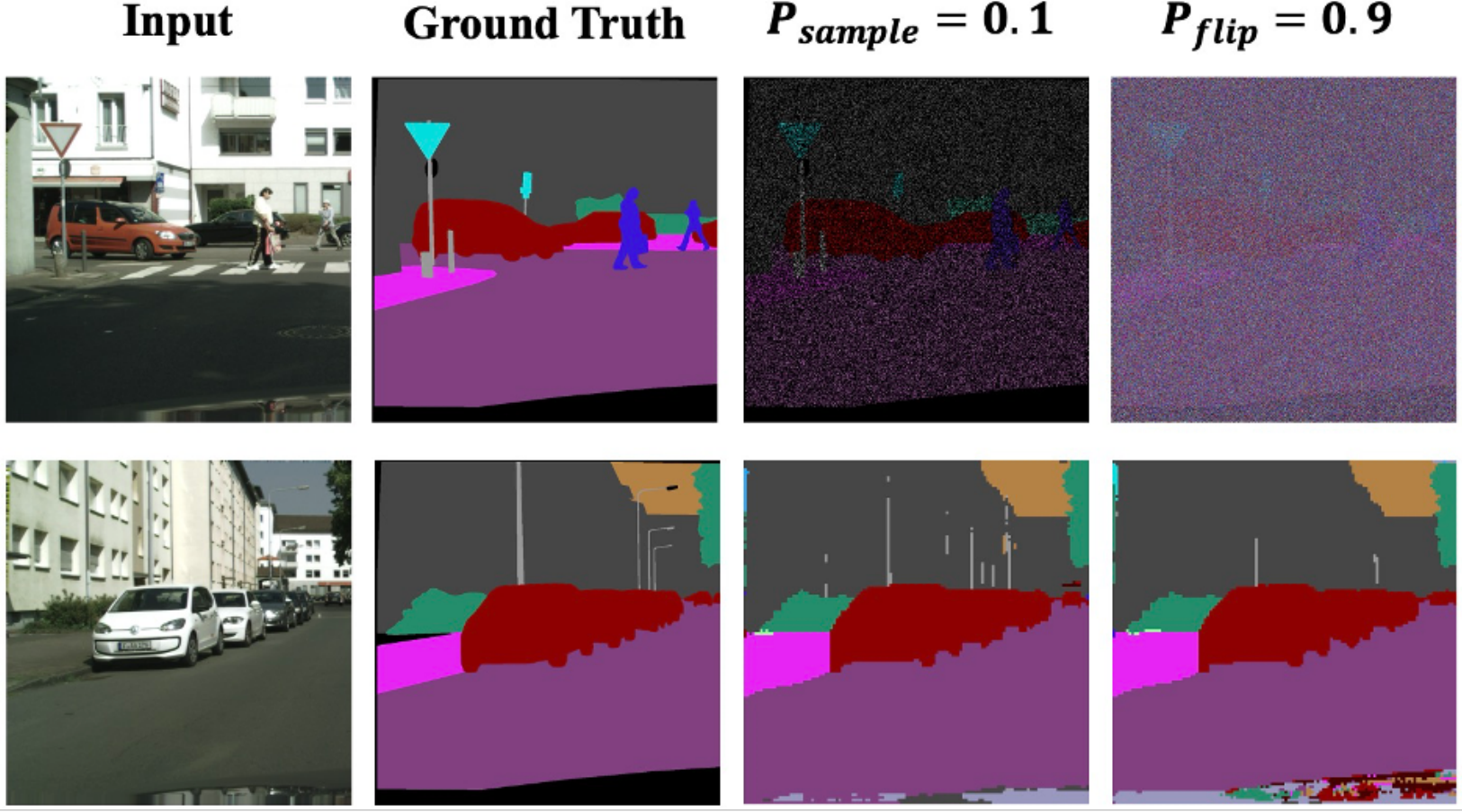} 
\caption{Segmentation performance of Deeplabv3 trained by noisy labels. Upper row: Examples of noisy-labels synthesized by random sampling($P_{sample}=0.1$) and random flipping($P_{flip}=0.9$). Lower row: Examples of segmented images in test set.}
\label{fig3}
\end{figure}

Taking the results together, we reason that semantic information contained in CL is completely or largely preserved in RCL.

\begin{table}[t]
\centering
\begin{tabular}{lccc}
\toprule
Noise Type & Noise Ratio & mIoU(\%) \\

\midrule

\multirow{1}{*} {None}      & 0        & \tabincell{c}{64.8 \\ (Chen, 2017)} \\ 
\cline{1-3}
   \multirow{3}{*}{RS}    	                  & 0.5		  & 64.8   \\
           								~ 	  & 0.3		  & 64.5   \\
           								~ 	  & 0.1 		  & 64.6	   \\ 	     
           								
\cline{1-3}
   \multirow{3}{*}{RF}                    	  & 0.5		  & 64.7   \\
           								~ 	  & 0.7		  & 64.6   \\
           								~ 	  & 0.9 		  & 64.7	   \\

\bottomrule
\end{tabular}
\caption{Segmentation performance of DeepLabv3 trained by randomly sampled (RS) and randomly flipped (RF) labels on Cityscapes.}
\label{table2}
\end{table}

\subsection{Meta-Structures vs. Pixel-level Labels}

So far, we have demonstrated that DNNs can learn segmentation from extremely noisy labels. However, it is unclear whether meta-structures or pixel-level labels contribute more to segmentation performance. In binary-class segmentation of fluorescence microscopy images, when $P_{flip}\leq0.49$, the fraction of correctly annotated pixels ($\geq$51\%) still exceeds the fraction of incorrectly annotated pixels ($\leq$49\%). This raises the possibility that DNNs learn from major correct pixel-level labels rather than meta-structures. 

To test this possibility, we generate entirely random labels, referred to as RL on ER dataset. Each pixel is randomly assigned to foreground with a certain generation probability. Sample image is shown in Figure 4B. Here, we set the generation probability as 0.1 and compare with the noisy label synthesized by random flipping with a probability $P_{flip} = 0.49$. While the randomly flipped labels still contain the meta-structures, the randomly generated labels do not. When we count the number of correctly annotated pixels using CL as the reference, we find that the pixel-level error rate of the randomly generated labels is around 31\%, which is much lower than the error rate of randomly flipped labels (49\%). If DNNs mainly learn from the pixel-level labels, the segmentation performance trained by RL would be better than the randomly flipped labels. 

However, segmentation performance of U-Net trained by RL is actually worse than randomly flipped labels (Figure 5A). Similar results using DeepLabv3+ are shown in Appendix C.4A. Together, these results further support that DNNs learn from meta-structures in labels rather than pixel-level labels per se in their supervised training for semantic segmentation.

\subsection{Summary}

Although RCL is uncommon in real-world applications, it allowed us to discover this counterintuitive learning behavior that DNNs learn meta-structure rather than pixel-level labels per se in segmentation. Furthermore, we find that:
\begin{center}
	\textit{DNNs trained by randomized labels that contain similar meta-structure information as the ground truth labels provide similar performance in semantic segmentation.}
\end{center}

Mathematical formulation and proof of this finding is presented later as Theorem 2 in the section \textit{Theoretical Analysis of Meta-Structures}.


\section{Further Characterization of Meta-Structures}

In this section, we further characterize the properties of meta-structures by analyzing PCL and RL.

\subsection{Performance of DNNs Trained by PCL}
 Inaccurate boundaries are a common source of label noise in image segmentation. We simulate inaccurate boundaries using dilation, erosion and skeleton of CL, which we refer to as PCL (Table 1). We examine the influence of PCL on segmentation performance and the results for U-Net are shown in Figure 5B. Results for DeepLabv3+ and more smooth perturbation on boundaries using different combinations of dilation and erosion are shown in Appendix C.4B-C.
 
Dilation and erosion of CL lead to degradation in segmentation accuracy (Figure 5B) because of their perturbation of meta-structures. Further degradation in segmentation performance under training by one-pixel-wide skeleton is because of the loss of width information. We conclude that:
 
\begin{center}
	\textit{Training of DNNs by labels with progressively stronger perturbation to the meta-structure exhibit progressively worse training performance}
\end{center}

Mathematical formulation and proof of this argument is presented later as Theorem 1 in the section \textit{Theoretical Analysis of Meta-Structures}.

\begin{figure*}[t]
\centering
\includegraphics[width=0.8\textwidth]{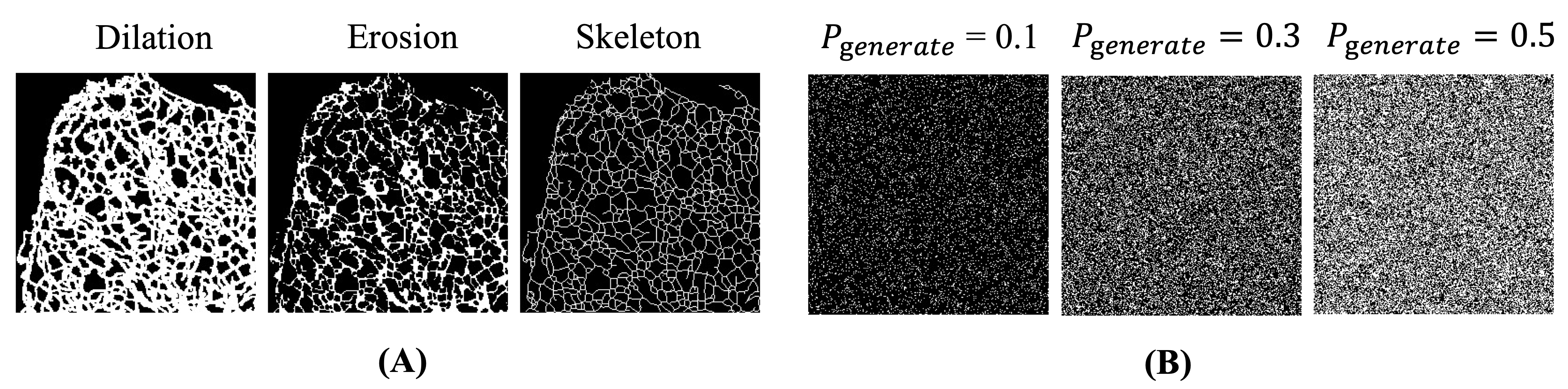} 
\caption{Different perturbations to ground truth (CL). (A): Examples of perturbed clean labels. (B): RL synthesized under different generation probabilities.}
\label{fig4}
\end{figure*}

\begin{figure*}[t]
\centering
\includegraphics[width=0.8\textwidth]{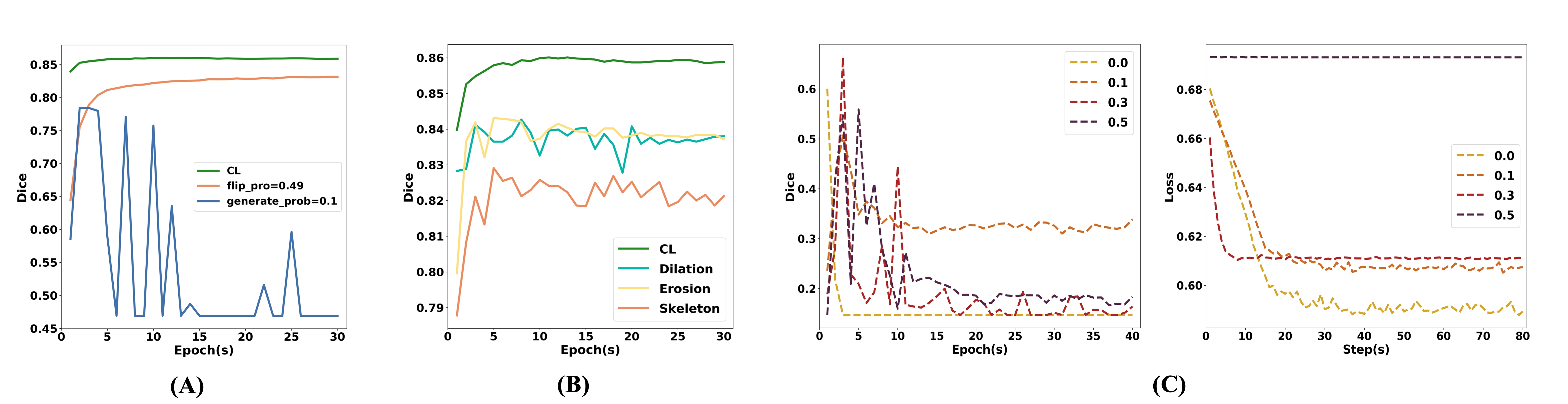} 
\caption{Performance of DNNs trained under different perturbations to ground truth (CL). (A): Testing dice score trained by CL, RCL ($P_{flip}=0.49$) and RL ($P_{generate}=0.1$). (B): Testing dice score trained by CL and PCL. (C): Testing dice scores (left) and training loss (right) trained by RL.}
\label{fig5}
\end{figure*}

\subsection{Learning behavior of DNNs Trained by RL}

 \citet{DBLP:conf/iclr/ZhangBHRV17} and \citet{DBLP:conf/icml/ArpitJBKBKMFCBL17} have shown that DNNs learn simple patterns before fitting RL by memorization. This conclusion, however, is drawn in image classification. In our study, we investigate whether DNNs exhibit similar behavior in image segmentation. We generate RL under different generation probabilities (Figure 4B) from 0 to 0.5. We experiment on the ER dataset and show the results of U-Net in Figure 5C. 

From the testing results on ER dataset (Figure 5C, left panel), we find that the learning process consists of two stages. In the first stage, the dice scores fluctuate substantially and reach several high values, indicating that the U-Net keeps learning and is not yet strongly influenced by RL. In the second stage, the dice scores drop quickly then converge to a low value, indicating that the U-Net start memorizing as the generalization ability becomes worse. Meanwhile, we find that under a higher generation probability, the training loss converges more quickly (Figure 5C, right panel), indicating DNNs have a higher tendency to memorize. Similar results are observed on the MITO dataset. See Appendix C.4D for further details. Overall, the learning behavior of DNNs trained by RL in segmentation is consistent with the learning behavior of DNNs trained by RL in classification.

\subsection{Summary}

Compared to CL and RCL, the PCL and RL contain different perturbations to meta-structures. We observe consistent degradation in segmentation performance.


\section{Unsupervised Binary-class Segmentation Based on Meta-Structures}

Here we propose an unsupervised method for binary-class segmentation of biological microscopy images, which we name as iterative ground truth training (iGTT). Our goal is to provide an example of utilizing meta-structures in practice.

\subsection{Notation}
Given data pair $(X,Y) \in \mathbb{R}^{H \times W}$, where $X$ denotes the input image, $Y$ denotes the binary-class labels with foreground and background labeled as 1 and 0, respectively. $H$ and $W$ denote image height and width, respectively. For our unsupervised method, we use the U-Net as the base model. The output of the final layer is denoted as $P$.

\subsection{Unsupervised Iteration Strategy}
We use fully black images to initialize the labels $Y^*$ then iteratively update it in following epochs. Specifically, for the $n_{th}$ epoch, the final layer of the image prediction score is $P^n=\mathcal{F}(\theta^{n}, X)$, where $\mathcal{F}$ denotes the model and $\theta$ are its parameters. We use sigmoid function in the final layer and only output the probability that belongs to the foreground for each pixel. We update the $\theta$ based on current epoch labels $Y^*$. Then we refresh $Y^*$ according to the $P^{n}$ by generating a threshold set $T$ with $K$ thresholds as follows:
\begin{equation}
	T=\{t\mid t = p_{min} + k \times \Delta \}
\end{equation}
\begin{equation}
	\Delta =  \frac{p_{max} - p_{min}}{K-1}
\end{equation}
where $k=\{0,1,\dots,K-1\}$. $p_{min}$ and $p_{max}$ denotes the minimum and maximum pixel value in $P^n$, respectively. $\Delta$ denotes the interval between neighboring thresholds in threshold set $T$. 

Based on thresholds $t$ in $T$, we generate $K$ coarse segmentation images $S_k^n$ ($k=0,1,\dots,K-1$) by thresholding $P^n$, where pixel $p$ is set to 1 if $p>t$. Next, we directly calculate the correlation between $P^n$ and $S_k^n$, then we find the most correlated segmentation image $\widetilde{S}^n$ from $S_{k\in\{0,\dots,K-1\}}^n$ and consider $\widetilde{S}^n$ as the optimal candidate labels for the next epoch of training.

Because we found most pixels $p_i^1$ of the output $P^1$ are close to 0 after first epoch training, if we used distance-based metrics for correlation calculation, the selected candidate labels $\widetilde{S}^n$ will tend to be black images since the distance of pair ($p_i^1$ , 0) is smaller than ($p_i^1$ , 1). To address this issue, we follow \cite{xu2019l_dmi} and select an information-theoretic noise-robust loss $\mathcal{L}_{DMI}$ to measure the correlation between $P^n$ and $S^n_k$ labels as follows:
\begin{equation}
	Cor(P^n,S_k^n) = \mathcal{L}_{DMI}(P^n,S_k^n) = -\log(\left| det(Q_{(P^n \parallel S_k^n)}) \right|)
\end{equation}
where $Q_{(P^n \parallel S_k^n)}$ is the matrix form of the joint distribution over $P^n$ and $S_k^n$. To calculate the $Q_{(P^n \parallel S_k^n)}$, we first resize the $P^{n} \in \mathbb{R}^{H \times W}$ to $P_f \in \mathbb{R}^{1 \times HW}$, then concatenate $P_f$ and $1-P_f$ to $\mathcal{P} \in \mathbb{R}^{2 \times HW}$. Meanwhile, we resize $S_k^n \in \mathbb{R}^{H \times W}$ to $S_f \in \mathbb{R}^{1 \times HW}$, then concatenate $S_f$ and $1-S_f$ to $\mathcal{S} \in \mathbb{R}^{2 \times HW}$. The $Q_{(P^n \parallel S_k^n)}$ is defined by matrix multiplication as follows:
\begin{equation}
	Q_{(P^n \parallel S_k^n)} = \mathcal{P} \mathcal{S}^{\mathrm{T}}
\end{equation}
By calculating $Cor(P^n,S_k^n)$, we find the optimal candidate labels $\widetilde{S}^n$ that has the highest mutual information with the prediction $P^n$. Then we send $\widetilde{S}^n$ into an EMS module (i.e., extraction-of-meta-structure module, see next section) to extract the meta-structures $S^{meta}$. Finally, we update the label $Y^*$ by $S^{meta}$. The whole iteration strategy is summarized in Algorithm 1. The architecture of iGTT is shown in Appendix D.2.

\begin{algorithm}[tb]
\caption{Unsupervised Iteration Strategy}
\label{iteration strategy}
\textbf{Inputs}: $X_{train} \in \mathbb{R}$ \\
\textbf{Parameter}: threshold $t$, model $\mathcal{F}$
\begin{algorithmic}[1] 
   \STATE $n=1$
   \STATE $Y^*=0$
   \IF{$n<Maxiters$}
   \STATE $P^n = \mathcal{F}(X_{train})$
   \STATE \bfseries UPDATE $\mathcal{F}$ with $Y^*$
   \FOR{$k \in \{0,1,...,K-1\}$}
   \STATE $ S_k^n\leftarrow\left\{
             \begin{array}{lr}
             p=1 $ \bfseries  if  $ p-t_k>0, p \in P^n \\
             p=0 $ \bfseries  if  $ p-t_k<0, p \in P^n
             \end{array} \right.$
   \ENDFOR
   \STATE $\widetilde{S}^n=argmin_{k \in \{0,...,K-1\}}Cor(P^n, S_k^n)$
   \STATE $Y^*=S^{meta}=EMS(\widetilde{S}^n)$
   \STATE $n=n+1$
   \ENDIF
\end{algorithmic}
\end{algorithm}

\subsection{Extraction-of-Meta-Structure Module}

Because of insufficient training, the segmented images $\widetilde{S}^n$ by thresholding are coarse in early training steps. Thus we do not directly use $\widetilde{S}^n$ as the next epoch training labels. However, the basic topology of objects are largely retained in $\widetilde{S}^n$. We have demonstrated that DNNs trained by labels with correct topology structures (meta-structures) can achieve similar performance as ground-truth labels. Based on this, we design an extraction-of-meta-structure (EMS) module to further improve the quality of the pseudo labels.

We first extract the skeleton of the $\widetilde{S}^n$, then we randomly shift every pixel in skeleton within a radius $r$ to disrupt the perturbation made by PCL. Since the randomly shift may move some pixel labels outside the target meta-structures, we follow a random sampling operation to filter out these pixel labels. The final pseudo label $S^{meta}$ generated by EMS module refines the meta-structures of the $\widetilde{S}^n$. Then we directly update $Y^*$ as the $S^{meta}$ and use $Y^*$ for the next epoch training.

To optimize the model, we combine $\mathcal{L}_{DMI}$ and $\mathcal{L}_{IOU}$ \cite{huang2019batching} to minimize the loss function $\mathcal{L}$ for current epoch:
\begin{equation}
\begin{aligned}
	\mathcal{L} = \underbrace{-\log (\left|det(\mathcal{P} \mathcal{S}^{\mathrm{T}})\right|)}_{\mathcal{L}_{DMI}} + \\ \underbrace{(1 - \frac{\sum_{i=1}^{HW}p_iy_{i}^*}
	{\sum_{i=1}^{HW}(p_i + y_{i}^* - p_iy_{i}^*) + \epsilon})}_{\mathcal{L}_{IOU}}
\end{aligned}
\end{equation}
where $p_i$ and $y_i^*$ denote the $i_{th}$ pixel in image $P$ and $Y^*$, respectively. $\epsilon$ is a smoothing coefficient to prevent the denominator from becoming zero.

\begin{table}[t]
\centering
\resizebox{.98\columnwidth}{!}{
\begin{tabular}{cclll}
\toprule
Dataset & Model & DICE(\%) & AUC(\%) & ACC(\%) \\

\midrule
\multirow{9}{*}{ER} & U-Net            & 85.99 & 97.09 & 91.09   \\
 & HRNet            & 86.07 & 97.17 & 91.18  \\
 & DeepLabv3+       & 81.66 & 94.80 & 87.67   \\
 \cline{2-5}
 & AGT			   & 76.23 & 82.63 & 85.19   \\
 & Otsu			   & 69.47 & 76.76 & 84.76  \\
 & DFC			   & 78.13 & 85.29 & 84.45  \\  
 & AC			   & 73.11 & 87.86 & 81.41   \\
 & iGTT(w EMS)      & $78.84_{1.17}$ & $91.61_{1.04}$ & $85.41_{1.06} $ \\
 & iGTT(w/o EMS) & $73.96_{0.97}$ & $84.53_{2.52}$ & $81.16_{1.03}$ \\

\bottomrule
\end{tabular}}
\caption{Segmentation performance of various models. Numbers in subscripts represent standard deviation.}
\label{table3}
\end{table}

\subsection{Segmentation Experiments}

As iGTT is customized for binary-class segmentation, we evaluate its performance on the ER, MITO \cite{t2he-zn97-20} and NUC \cite{caicedo2019nucleus} datasets. Refer to Appendix D.1 for further configuration details. 

To compare with supervised methods, we select two commonly used DNNs (U-Net, DeepLabv3+) and a state-of-the-art model HRNet \cite{wang2020deep}. To compare with unsupervised methods, we select adaptive gaussian thresholding (AGT), Otsu, and two state-of-the-arts methods which including Autoregressive Clustering (AC) \cite{ouali2020autoregressive} and Differentiable Feature Clustering (DFC) \cite{kim2020unsupervised}. We use DICE(dice scores), AUC(area under curve) and ACC(accuracy) as the performance metrics. To reduce the effects brought by randomization, we trained our model 10 times and calculate the mean and standard deviation. 

Segmentation results on ER are summarized in Table 3. See Appendix D.3 for results on MITO and NUC datasets. For supervised models, HRNet achieves the best performance. The DeepLabv3+ performs worse than U-Net and HRNet. For unsupervised methods, iGTT achieves the best performance. Meanwhile, we find that iGTT achieves competitive performance when comparing with the other three supervise models. Moreover, we find that using EMS module improves the final segmentation performance, indicating that EMS indeed refines the candidate labels. Overall, our model effectively narrows the gap between supervised learning and unsupervised learning by effectively utilizing the implicit meta-structures in noisy-labels. Examples of testing segmented images on all datasets are shown in Appendix D.4 and D.5.


\section{Theoretical Analysis of Meta-Structures}
We model meta-structures of labels based on the theorem of spatial point analysis \cite{baddeley2015spatial,diggle2013statistical,illian2008statistical}. Specifically, we use $P(y^*=j|y=i)$ to denote the probability of flipping ground-truth pixel label $y$ in class $i \in \{1,\dots,M \}$ to the noisy pixel label $y^*$ in class $j \in \{1,\dots,M \}$, where $M$ denotes the number of semantic classes. For RCL and RL, we can build noise transition matrix $Q_{y^*|y}$ based on $P(y^*=j|y=i)$. Examples of $Q_{y^*|y}$ are shown Figure 6. 


For binary-class semantic labels, we treat the foreground pixels ($y^*=1$) as the spatial data point. For multiple-class semantic labels, we disassemble multi-class labels as multiple binary-class labels and separately analysis the spatial density distributions of each class. Specifically, We treat the pixel $x^m$ whose label $y^*=m \in \{1,\dots,M \}$ as the spatial data point and view other pixels $(y^* \neq m)$ as background by annotating as 0. We use $N$ to denote the number of spatial data points, $h$ to denote the bandwith (radius) of the search area $S$, and $K$ to denote the kernel function.

\textbf{Definition.} We define the Meta-structures of a label (MS) as a set of sematic classes $MS=\{O_1,...,O_m\}$, where each class $O_i=\{x^m|x^m \sim f_i (x^m) \}$ is composed of pixels $x^m$ that are drawn from the same underlying spatial density distribution $f_i (x^m)$.

\textbf{Lemma 1.} When pixels $x^m$ are treated as data points, the density distribution $f_i(x^m)$ in random noisy labels can be calculated as follows: 
\begin{equation}
	f_i(x^m)=\{ \frac{1}{2Nh} * \sum_{j=1}^M P(y^*=m|y=j)*S_j \} \pm \delta
\end{equation}
where $\delta$ denotes sampling errors, which are a constant, $S_j=S \cap O_j$ is the area that corresponds to $j_{th}$ semantic class $O_j$ within the search area $S$.

\textbf{Proof:} Based on the theorem of spatial point analysis, we use kernel function $K$ to estimate the density distribution $f_i(x^m)$ by counting $x^m$ within a search area as follows:
\begin{equation}
	f_i(x^m)= \frac{1}{2Nh} \sum_{k=1}^N K(x-h \le x_k^m \le x+h)
\end{equation}
Note that for RCL, the number of counted data points within the search area can be formulated as follows: 
\begin{equation}
	\sum_{k=1}^N K(x-h \le x_k^m \le x+h) = \sum_{j=1}^M P(y^*=m|y=j)*S_j \pm \delta
\end{equation} 
Because $N$, $h$, $\delta$ and $S_j$ are all constant, $f_i(x^m)$ is only dependent on flipping probability $P(y^*=m|y=j)$. \textit{QED} (See further details in Appendix B.1)

\textbf{Lemma 2.} The number of semantic classes $D$ in random labels equals the rank $R$ of noise transition matrix $Q_{y^*|y}$: $D=R$.

\textbf{Proof:}  If $R<M$, there exits $(M-R+1)$ linearly correlative columns and $P(y^*=j|y=1)=...=P(y^*=j|y=M-R)$. Based on Lemma 1, $f_i(x^m)$ is almost the same constant $C$ within the area $A=\{y=1\} \cup ... \cup \{y=M-R\}$, indicating that $(M-R+1)$ density distributions of the area $A$ are fused into one. Based on the definition that semantic class is composed of pixels that have similar density distributions, the number of semantic classes $D=M-(M-R)=R$.

If $R=M$, for each column of $Q_{y*|y}$ there at least exists $P(y^*=m|y=p) \neq P(y^*=m|y=q)$ and the $f_i(x^m)$ are different within the area $A=\{y=p\}\cup\{y=q\}$. Thus distribution differences on all classes exist and the number of semantic classes $D=M=R$. In summary, $D=R$. \textit{QED}

\textbf{Theorem 1.} For PCL, more bias on boundary pixels of ground-truth more perturbation on meta-structures.

\textbf{Proof:} According to the definition of meta-structure, the semantic boundaries between $Y^*$ and $Y$ are different since more bias on boundary pixels in PCL will lead to less interaction of semantic class $O_i$ between PCL and CL, and therefore more perturbation on meta-structures. \textit{QED} 

\textbf{Theorem 2.} If RCL $Y^*$ is synthesized by $Q_{y^*|y}$ and the rank of $Q_{y^*|y}$ is full, the noisy labels $Y^*$ have the similar semantic information as ground-truth $Y$.

\textbf{Proof:} Based on Lemma 2, if $R(Q_{y^*|y})=M$, the meta-structure of $Y^*$ contains the complete semantic classes as $Y$: $MS(Y^*) = MS(Y)= \{ O_1,\dots,O_m \}$. Furthermore, the location of difference between density distributions is unchanged, indicating semantic areas of different classes are also unchanged. Thus, all classes in $Y^*$ keep distinguishability like $Y$ and both of them contain similar semantic information. \textit{QED}

 We also demonstrate Theorem 2 by experiments. We generate a simplified binary image of (256×256) that has a circle within a rectangle as shown in Figure 6 (first column). For the noisy labels (RCL) whose rank equals 2 (second column), its density distribution estimated by kernel function exhibits similar patterns as the ground-truth, indicating they have similar meta-structures. However, for the noisy labels (RL) whose rank equals 1 (third column), its density distribution is randomized, indicating that two semantic classes are fused together. More experiment results are shown in Appendix B.2.
 

\textbf{Summary.} Here we provide a direct mathematical definition of meta-structures. This definition theoretically and experimentally explains the key observed results in our segmentation experiments.  

 \begin{figure}[t]
\centering
\includegraphics[width=0.9\columnwidth]{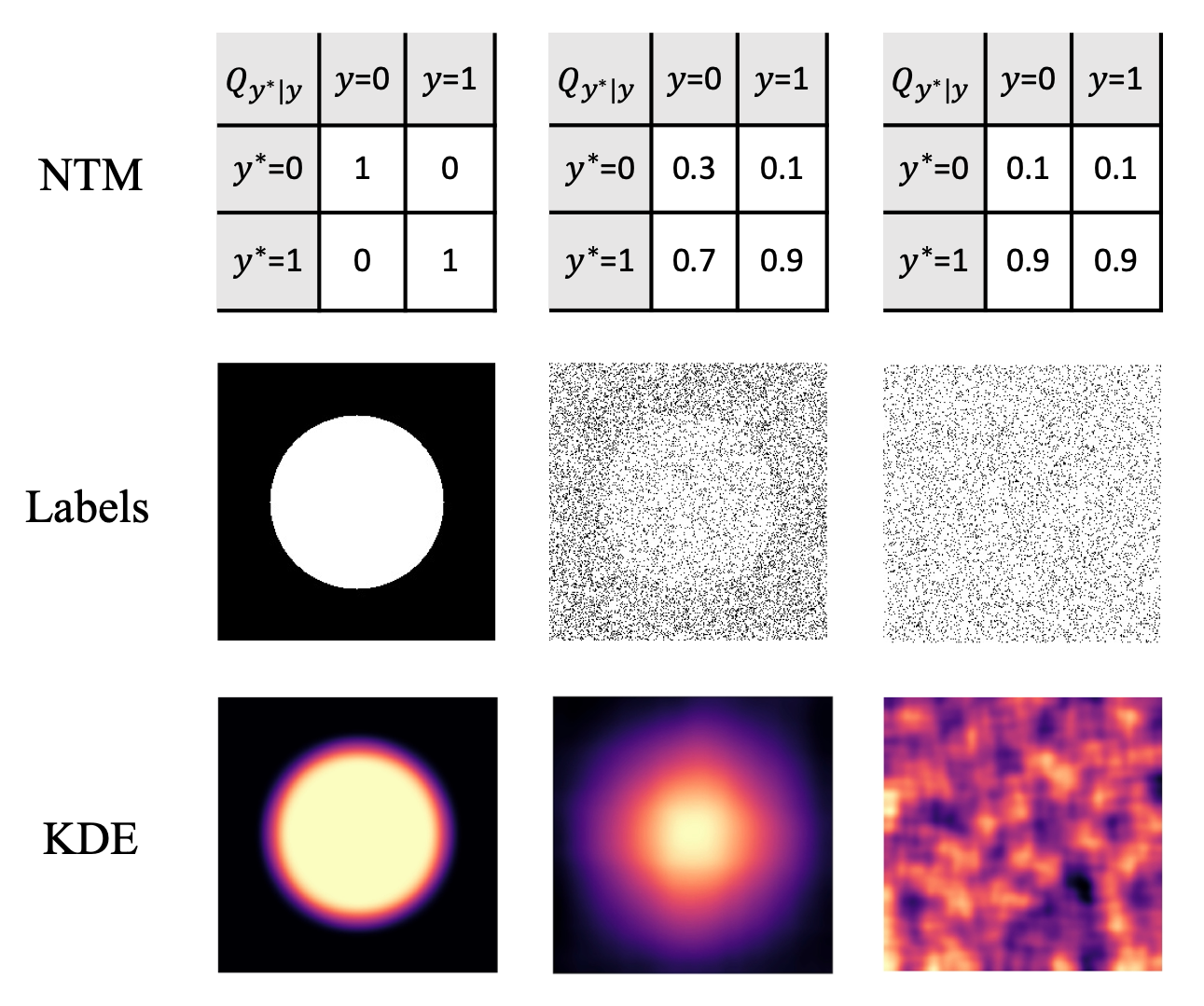} 
\caption{Estimation of density distribution of noisy labels with different ranks (from left to right: ground truth, RCL, RL). NTM: noise transition matrix. KDE: kernel density estimation.}
\label{fig6}
\end{figure}

\section{Related Work}

\textbf{Training with noisy labels.} To explore the generalization properties of DNNs trained with noisy labels, \citet{DBLP:conf/iclr/ZhangBHRV17} and \citet{DBLP:conf/icml/ArpitJBKBKMFCBL17} performed a series of experiments and demonstrated that DNNs can easily memorize random labels but with poor generalization. This phenomenon contradicts traditional statistical learning theory \cite{vapnik1999overview, bartlett2005local} and has attracted a large number of studies on how to mitigate the negative influence of noisy labels in deep learning.

In image classification, many studies have tried to propose noise-robust loss functions \cite{manwani2013noise, masnadi2008design, brooks2011support, van2015learning, ghosh2017robust, zhang2018generalized, xu2019l_dmi}. Several works focus on designing custom architectures for deep neural networks \cite{jiang2018mentornet, han2018co}. Transfer learning has also been applied \cite{lee2018cleannet}. Some other studies proposed multi-tasks frameworks to estimates true labels \cite{veit2017learning, tanaka2018joint, yi2019probabilistic, li2020dividemix}. Although most of these methods performed well in image classification tasks, their performance in image segmentation remains unknown.

In semantic segmentation, \citet{min2019two} proposed to weaken the influence of back-propagated gradients caused by incorrect labels based on mutual attention. \citet{shu2019lvc} proposed to leverage local visual cues to automatically correct label errors. Several studies also proposed semi-supervised or unsupervised methods \cite{lu2016learning, li2019supervised, navlakha2013unsupervised, zheng2020rectifying}.

All these studies focus on improving the generalization ability of DNNs trained with noisy labels. However, an in-depth understanding of the fundamental roles of labels in training remains lacking. We address this deficiency by focusing on the noisy labels and examine their influence on image segmentation directly and systemically.  

\textbf{Unsupervised segmentation.} Some works try to segment images by learning pixel representation in a self-supervised setting \cite{hwang2019segsort, zhang2020self}. However, they still relied on initialization from other annotated datasets. A limited number of studies address image segmentation in a fully unsupervised way. Methods have been proposed to maximize the mutual information between augmented views \cite{ouali2020autoregressive, ji2019invariant}. A DNN architecture consisting of convolutional filters for feature extraction and differentiable processes for feature clustering has also been proposed \cite{kim2020unsupervised}. Overall, performance of unsupervised segmentation methods remains to be further improved.


\section{Conclusion}

In this study, we examine the learning behavior of DNNs trained by different types of pixel-level noisy labels in semantic segmentation and provide direct experimental evidence and theoretical proof that DNNs learn meta-structures from noisy labels. The unsupervised segmentation model we have developed provides an example on how to utilize the meta-structures in practice. However, our study also has its limitations. In particular, our model of meta-structures remains to be further developed, and utilization of meta-structures remains to be expanded to other applications such as multi-class segmentation. Despite these limitations, the learning behavior of DNNs revealed in this study provides new insight into what and how DNNs learn from noisy labels to segment images.

\appendix

\section{Acknowledgments}
The authors thank members of CBMI for their technical assistance. The work was supported in part by the National Natural Science Foundation of China (grant 31971289 and 91954201 to G.Y.) and the Strategic Priority Research Program of the Chinese Academy of Sciences (grant XDB37040402 to G.Y.).


\nocite{li2019supervised}
\nocite{lu2016learning}
\nocite{long2015fully}
\nocite{jegou2017one}
\nocite{chen2017deeplab}
\nocite{zheng2020rectifying}
\nocite{DBLP:journals/corr/YuK15}
\nocite{ronneberger2015u}
\nocite{chen2018encoder}
\nocite{shu2019lvc}
\nocite{he2016deep}
\nocite{huang2019batching}
\nocite{caicedo2019nucleus}
\nocite{t2he-zn97-20}
\nocite{bartlett2005local}
\nocite{min2019two}
\nocite{vapnik1999overview}
\nocite{DBLP:conf/iclr/ZhangBHRV17}
\nocite{DBLP:conf/icml/ArpitJBKBKMFCBL17}
\nocite{xu2019l_dmi}
\nocite{manwani2013noise}
\nocite{masnadi2008design}
\nocite{brooks2011support}
\nocite{van2015learning}
\nocite{ghosh2017robust}
\nocite{zhang2018generalized}
\nocite{li2020dividemix}
\nocite{navlakha2013unsupervised}
\nocite{jiang2018mentornet}
\nocite{han2018co}
\nocite{lee2018cleannet}
\nocite{tanaka2018joint}
\nocite{veit2017learning}
\nocite{yi2019probabilistic}
\nocite{wang2020deep}
\nocite{ouali2020autoregressive}
\nocite{kim2020unsupervised}
\nocite{ji2019invariant}
\nocite{hwang2019segsort}
\nocite{zhang2020self}
\nocite{cordts2016cityscapes}
\nocite{baddeley2015spatial}
\nocite{diggle2013statistical}
\nocite{illian2008statistical}

\bibliography{aaai22.bib}

%
%
%
%


%
%

\end{document}